\documentclass[format=acmsmall,nonacm,review=false, screen=true]{acmart}

\usepackage{pgfplotstable}
\usepackage{amsmath,amsfonts}
\usepackage{array}
\usepackage{colortbl}
\usepackage{booktabs} 
\usepackage{pbox}
\usepackage{mathtools}
\usepackage{makecell}  

\usepackage{xcolor}

\usepackage{wrapfig}
\usepackage{lscape}   
\usepackage{longtable}
\usepackage{tabu}
\usepackage{multicol}
\usepackage{array} 
\usepackage[T1]{fontenc}
\usepackage{tikz}

\usepackage{graphicx}
\usepackage{siunitx}
\usepackage{pgfplotstable}

\usepackage[ruled]{algorithm2e} 

\SetAlFnt{\small}
\SetAlCapFnt{\small}
\SetAlCapNameFnt{\small}
\SetAlCapHSkip{0pt}
\IncMargin{-\parindent}

\acmJournal{TWEB}
\acmVolume{9}
\acmNumber{4}
\acmArticle{0}
\acmYear{2010}
\acmMonth{3}
\copyrightyear{2009}
\acmArticleSeq{9}

\setcopyright{none}
\renewcommand\footnotetextcopyrightpermission[1]{}
\begin{document}
\title{The Tribes of Machine Learning and the Realm of Computer Architecture}
\author{Ayaz Akram}
\affiliation{%
  \institution{University of California, Davis}
}
\email{yazakram@ucdavis.edu}
\author{Jason Lowe-Power}
\affiliation{%
  \institution{University of California, Davis}
 }
\email{jlowepower@ucdavis.edu}

\begin{abstract}
Machine learning techniques have influenced the field of computer architecture like many other fields.
This paper studies how the fundamental machine learning techniques can be applied towards computer architecture problems.
We also provide a detailed survey of computer architecture research that employs different machine learning methods.
Finally, we present some future opportunities and the outstanding challenges that need to be overcome to exploit full potential of machine learning for computer architecture.
\end{abstract}

%
%


%
%


\maketitle
\thispagestyle{empty}

\renewcommand{\shortauthors}{Akram and Lowe-Power}

\section{Introduction}

Machine learning (ML) refers to the process in which computers learn to make decisions based on the given data set without being explicitly programmed to do so~\cite{alpaydin2014introduction}.
There are various classifications of the ML algorithms.
One of the more insightful classifications has been done by Pedro Domingos in his book \textit{The Master Algorithm} \cite{domingos2015master}.
Domingos presents five fundamental tribes of ML: the symbolists, the connectionists, the evolutionaries, the bayesians and the analogizers.
Each of these believe in a different strategy to go through the learning process.
These tribes or schools of thought of ML along-with their primary algorithms and origins are shown in Table~\ref{tab:tribes}.
There are existing proofs that given the enough amount of data, each of these algorithms can fundamentally learn anything.
Most of the well known ML techniques/algorithms~\footnote{See Appendix~\ref{sec:overview} for a detailed summary of many ML techniques referred in this paper.} belong to one of these tribes of ML.

\begin{table}[h!]
  \begin{center}
    \caption{Five Tribes of ML (taken from \cite{domingos2015master})}
    \label{tab:tribes}
    \begin{tabular}{|l|l|r|} \Xhline{2\arrayrulewidth} 
     \rowcolor{brown!25}
     \textbf{Tribe} & \textbf{Origins} & \textbf{Master Algorithms}\\ \Xhline{2\arrayrulewidth}
     \rowcolor{brown!25}
      Symbolists & Logic, philosophy & Inverse deduction \\ \hline
     \rowcolor{brown!25}
      Connectionists & Neuroscience & Backpropagation \\ \hline
     \rowcolor{brown!25}
      Evolutionaries & Evolutionary biology & Genetic programming \\ \hline
     \rowcolor{brown!25}
      Bayesians & Statistics & Probabilistic infernce \\ \hline
     \rowcolor{brown!25}
      Analogizers & Psychology & Kernel machines \\ \Xhline{2\arrayrulewidth}
    \end{tabular}
  \end{center}
\end{table}

In this paper, we look at these five school of thoughts of ML and identify how each of them can be fundamentally used
to solve different research problems related to computer architecture.
ML techniques have already influenced many domains of computer architecture.
Figure 1 shows the number of research works using ML each year since 1995.
It is observed that most of the work employing ML techniques (approximately 65\% of the studied work) is done in last 5-6 years
indicating increasing popularity of ML models in computer architecture research.
Findings also indicate that Neural Networks are the most used ML technique in computer architecture research as shown in Figure 2.

  \pgfplotstableread[col sep=comma]{data2.csv}{\datatable}
  \pgfplotstabletranspose[colnames from=Year, input colnames to=Works]{\transposed}{\datatable}

  \begin{figure}
  \centering
  \begin{tikzpicture}[scale=0.9]
  \begin{axis}[
      ybar,
      xtick=data,
      xticklabels from table={\datatable}{[index]0},
      xticklabel style={rotate=90},
      enlarge y limits=upper,
      ymin=0,
      bar width=5.5,
      ylabel=Number of Works
  ]
  \addplot [fill=cyan!70, draw=cyan!50!black]  table [
      x expr=\coordindex,
      y index=1] {\datatable};
  \end{axis}
  \end{tikzpicture}
  \caption{Number of works done in chronological order}
  \end{figure}
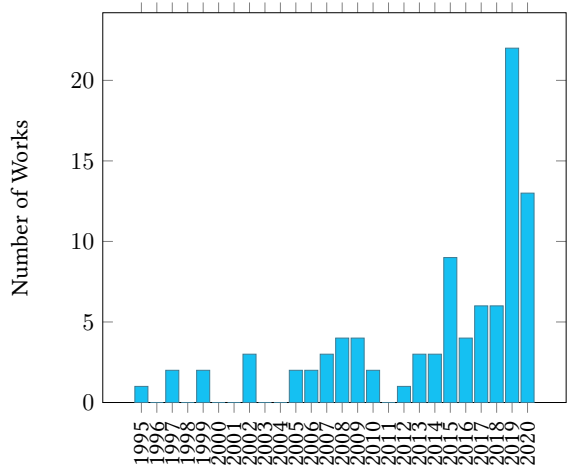

\pgfplotstableread[col sep=comma]{data.csv}{\datatable}
\pgfplotstabletranspose[colnames from=Algorithm, input colnames to=Works]{\transposed}{\datatable}

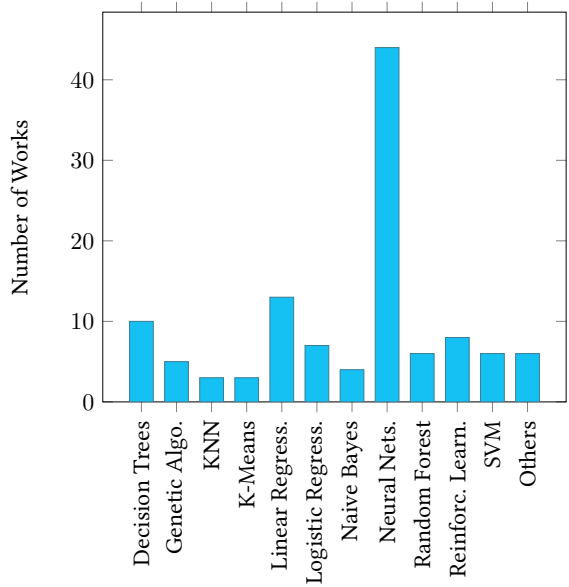
\begin{figure}
\centering
\begin{tikzpicture}[scale=0.9]
\begin{axis}[
    ybar,
    xtick=data,
    xticklabels from table={\datatable}{[index]0},
    xticklabel style={rotate=90},
    enlarge y limits=upper,
    ymin=0,
    ylabel=Number of Works
]
\addplot [fill=cyan!70, draw=cyan!50!black]  table [
    x expr=\coordindex,
    y index=1] {\datatable};
\end{axis}
\end{tikzpicture}
\caption{Number of works for each ML Algorithm}
\end{figure}

We also present an extensive survey of the research works employing these techniques in computer architecture.
Most of the insights discussed in this paper are based on a review of more than a hundred papers \footnote{Detailed summaries of most of these papers are available in: \url{https://github.com/ayaz91/Literature-Review/blob/main/ML\_In\_CompArch/ML\_CompArch.md}} which use ML for computer architecture.

The specific questions we answer in this paper include:

\begin{itemize}
\item
What are the fundamental features of different ML algorithms which make them suitable for
particular architecture problems compared to the others?
\item How has ML impacted computer architecture research so far?
\item What are the most important challenges that need to be addressed to fully utilize ML potential in computer architecture research?
\end{itemize}

\section{Survey of use of Machine Learning in Computer Architecture}

\begin{table*}[h!]
  \begin{center}
    \caption{ML Tribes and Examples of Their Use in Computer Architecture}
    \label{tab:survey}
    \begin{tabular}{|m{2.3cm}|m{1.5cm}|m{1.5cm}|m{7cm}|} \Xhline{2\arrayrulewidth}  
     \rowcolor{brown!25}
     \textbf{Tribe} & \textbf{Targeted Problem} & \textbf{Example Algorithms} & \textbf{Example Works}\\ \Xhline{2\arrayrulewidth}
      \rowcolor{brown!25}
      \textbf{Symbolists} & Knowledge Composition & Inverse deduction, Decision trees & Microarchitecture \cite{fern2000dynamic,liao2009machine,rahman2015maximizing,navarro2017machine}, Performance Estimation \cite{elmoustapha2007comparison,jundt2015compute}, Scheduling \cite{helmy2015machine}, Energy \cite{moeng2010applying}, instruction scheduling \cite{monsifrot2002machine,malik2008learning}\\ \hline
      \rowcolor{brown!25}
      \textbf{Connectionists} & Credit Assignment & Deep-Neural Networks, Perceptrons & Microarchitecture \cite{calder1997evidence,jimenez2002neural,wang2005dynamic,jimenez2003fast,seznec2014tage,mao2017study,teran2016perceptron,dai2016block2vec,khakhaeng2016finding,yazdanbakhsh2015neural}, Performance Estimation \cite{elmoustapha2007comparison,wu2015gpgpu}, scheduling \cite{bitirgen2008coordinated,li2009machine,wang2009mapping,wang2014integrating,helmy2015machine,nemirovsky2017machine}, DSE \cite{ipek2006efficiently,khan2007using}, Energy  \cite{won2014up}, Security \cite{ozsoy2016hardware,patel2017analyzing,khasawneh2018ensemblehmd,khasawneh2017rhmd,chiappetta2016real} \\ \hline
    \rowcolor{brown!25}
     \textbf{Evolutionaries} & Structure Discovery & Genetic Algorithm & Performance estimation \cite{hoste2006performance} , Design Space Exploration \cite{emer1997language, stanley1995parallel,reagen2017case}, instruction scheduling \cite{cooper1999optimizing,leather2009automatic} \\ \hline
     \rowcolor{brown!25}
     \textbf{Bayesians} & Uncertainty Reduction & Naive-Bayes, LDA & Microarchitecture \cite{beckmann2017maximizing}, DSE \cite{reagen2017case} \\ \hline
     \rowcolor{brown!25}
     \textbf{Analogizers} & Similarity Discovery & SVM, KNN & Microarchitecture \cite{culpepper2005svms,liao2009machine}, Performance \cite{elmoustapha2007comparison,baldini2014predicting,hoste2006performance}, Scheduling \cite{wang2009mapping,wang2014integrating,helmy2015machine}, Security \cite{patel2017analyzing} \\ \Xhline{2\arrayrulewidth}
     \rowcolor{brown!25}
    \end{tabular}
  \end{center}
\end{table*}




In this section, we will discuss how each of the previously mentioned paradigms of ML can be (or have been) applied to the field of computer architecture.
Table \ref{tab:survey} enlists the five tribes of ML along-with their targeted problem and some examples of the relevant research works.
Of course, many architectural problems can be solved by more than one of these families of ML algorithms.

\subsection{The Symbolists:}
This tribe of ML relies on symbol manipulation to produce intelligent algorithms.
The fundamental problem that Symbolists are trying to solve is knowledge composition.
Their insight is to use initial knowledge to learn quicker than learning from scratch.
More specific examples of this group of ML algorithms include inverse deduction and decision trees.
Algorithms belonging to Symbolists seem to be ideal to be used for cases where a cause and effect relationship needs to be established between events.
For example, the architecture problems where we need this type of learning include: finding reasons for hardware security flaws and consistency bugs.

The more interpretable nature of these algorithms make them a good candidate to understand the impact of certain design features (input events) on the performance metric of the entire system.
As a result, certain parameters can be fine tuned to produce desirable results.
For example, this kind of learning can be applied to understand which pipeline structures consume the most amount of energy in a given configuration.
However, it might be hard to map these algorithms directly to the hardware (if desired to be used at run-time) due to their symbolic nature.

There exist some examples of usage of these algorithms in computer architecture research.
Fern et al. \cite{fern2000dynamic} introduced decision tree based branch predictors.
Decision trees allowed the proposed branch predictor to be controlled by different processor state features.
The relevant features could change at run-time without increasing linearly in size with the addition of new features (compared to table based predictors), providing significant benefits over conventional table based predictors.

Decision tree based models have been used in many cases to understand the impact of different architectural events on systems' performance.
~Yount et al. \cite{elmoustapha2007comparison} compared various machine learning (ML) algorithms with respect to their ability to analyze architectural performance of different workloads and found tree-based ML models to be the most interpretable and almost as accurate as Artificial Neural Networks (ANNs).
Jundt et al. \cite{jundt2015compute} used a ML model named Cubist \cite{Cubist}, which uses a tree of linear regression models to observe the importance
of architectural blocks that have an effect on performance and power of high performance computing applications on Intel's Sandy Bridge and ARMv8 XGene processors.
Mariani et al. \cite{mariani2017predicting} used Random Forest to estimate HPC (High Performance Computing) applications' performance on cloud systems using hardware independent profiling of applications. 

~Rahman et al. \cite{rahman2015maximizing} used decision trees and logistic regression to build framework to identify the best prefetcher configuration for given multithreaded code (in contrast to focus on serial code as in~\cite{liao2009machine}). Hardware prefetcher configuration guided by the presented machine learning framework achieved close to 96\% speed-up of optimum configuration speed-up.

Moeng and Melhem \cite{moeng2010applying} used decision trees (implemented in hardware) to propose a DVFS (dynamic voltage and frequency scaling) policy that will be able to predict clock frequency resulting into the least amount of energy consumption in a multicore processor.
They used simple measurement metrics like cycles, user instructions, total instructions, L1 accesses and misses, L2 accesses, misses and stalls for a DVFS interval during execution as inputs to the decision trees.

\subsection{The Connectionists:}
The connectionists rely on brain structure to develop their learning algorithms and try to learn connections between different building blocks (neurons) of the brain.
The main problem they are trying to solve is of credit assignment i.e. figure out which connections are responsible for errors and what should be the actual strength of those connections.
The most common example of this class of ML algorithms is (deep) neural networks.
The algorithms belonging to the Connectionists are good at learning complex patterns specially at run time (as evident by many microarchitecture design examples).
This type of learning can be useful in many microarchitectural predictive structures (like branch predictors, cache prefetchers and value predictors) which try to forecast an event based on similar events of the past, where different past events might have different weights in determining if a certain event will happen in the future (the problem of credit assignment).
In contrast to symbolists, the algorithms belonging to this group might not need any initial knowledge, however they are not very interpretable.
So, they might not be very useful to understand the importance of different architectural events/components in overall performance/output for static analysis.
This kind of algorithms have found uses in the architecture research, specially simple neural networks like perceptrons \cite{block1962perceptron} (owing to their simple structure and being more amenable to be implemented in the hardware).

Calder et al. \cite{calder1997evidence} introduced the use of neural networks for static branch prediction in the late 1990's.
One of the earliest works to use ML for dynamic branch prediction was done by Vinton and Iridon \cite{vintan1999towards}. They used neural networks with Learning Vector Quantization (LVQ) \cite{kohonen1995learning} as a learning algorithm for neural networks and were able to achieve around 3\% improvement in misprediction rate compared to conventional table based branch predictors.
Later, Egan et al. \cite{egan2003two} and Wang and Chen \cite{wang2005dynamic} also used LVQ for branch prediction.
The complex hardware implementation of LVQ due to computations involving floating point numbers, could significantly increase the latency of the predictor~\cite{jimenez2002neural}.
Jimenez et al.~\cite{jimenez2002neural}, working independently, also used neural network based components, perceptrons \cite{block1962perceptron}, to perform dynamic branch prediction.
In their work, each single branch is allocated a perceptron.
The inputs to the perceptron are the weighted bits of the "global branch history shift register", and the output is the decision about branch direction.
One big advantage of perceptron predictor is the fact that the size of perceptron grows linearly with the branch history size (input to the perceptron) in contrast the size of pattern history table (PHT) in PHT based branch predictors which grows exponentially with the size of the branch history.
Therefore, within same hardware budget, perceptron based predictor is able to get a benefit from longer branch history register.
One of the big problems with the use of perceptrons is their inability to learn linear separability \footnote{A boolean function is "linearly separable" if all false instances of the function can be separated from its all true instances using a hyperplane \cite{mittal2018survey}. As an example XOR is linearly inseparable and AND is linearly separable.}.
This was later resolved by Jimenez \cite{jimenez2005piecewise} using a set of piecewise linear functions to predict the outcomes for a single branch. These linear functions refer to a distinct historical path that lead to the particular branch instruction. Graphically, all these functions, when combined together, form a surface.


Since the introduction of neural network based branch predictors, there has been a lot of research done to optimize the design of these predictors. Different analog versions have been proposed to improve speed and power consumption of the neural network based branch predictors \cite{st2008low,jimenez2011optimized,amant2009mixed,saadeldeen2013memristors,kirby2007mixed}. Perceptron based predictors have also been combined with other perceptron or non-neural network based predictors to achieve better accuracy overall: \cite{tarjan2005merging,egan2003two,monchiero2005combined,jimenez2009generalizing,srinivasan2007idealistic,ribas2006evaluating,lee2009multiple,jimenez2016multiperspective}. Different optimizations/modifications to neural network branch predictors including use of different types of neural networks have been explored: \cite{jimenez2003fast,gao2005adaptive,seznec2004revisiting,cadenas2005new,ninomiya2006path,kim2003branch,sethuram2007neural,ho2007combining,gope2014bias,tu2007hardware,shiperceptron,gaudet2016using,jimenez2005improved}. These optimizations improve performance of the predictor and save power and hardware cost. Similar concepts are also used in development of other branch predictor related structures like confidence estimators and indirect branch predictors: \cite{akkary2004perceptron,jimenez2001perceptron,desmet2006improved,kelleybranch,jimenezsnip}.
The statistical corrector predictor used in TAGE-SC\_L \cite{seznec2014tage} branch predictor~\footnote{ won the 2014 branch predictor championship and combines TAGE branch predictor with a loop predictor and a statistical corrector predictor} is also a perceptron based predictor.
This statistical corrector predictor regresses TAGE's prediction if it statistically mispredicts in similar situations.
Mao et al. \cite{mao2017study}, recently, applied deep learning (easy to train Deep Belief Networks (DBN)  \cite{hinton2007learning} to the problem of branch prediction.


The popularity of perceptrons in branch prediction has also affected the design of other microarchitecture structures.
For example, Wang and Luo \cite{wang2017data} proposed perceptron based data cache prefetching.
The proposed prefetcher is a two level prefetcher which uses conventional table-based prefetcher at the first level.
At the second level a perceptron is used to reduce unnecessary prefetches by relying on memory access patterns. 
Peled et al. \cite{peled2018towards} used neural networks to capture semantic locality of programs.
Memory access streams alongwith a machine state is used to train neural network at run-time which predicts the future memory accesses.
Evaluation of the proposed neural prefetcher using SPEC2006 \cite{spec2006Web}, Graph500 \cite{murphy2010introducing} benchmarks and other hand-written kernels indicated an average speed-up of 22\% on SPEC2006 benchmarks and 5x on other kernels.
However, importantly, Peled et al. \cite{peled2018towards} also performed a feasibility analysis of the proposed prefetcher which shows that the benefits of neural prefetcher are outweighed by other factors like learning overhead, power and area efficiency.
Teran et al. \cite{teran2016perceptron} applied perceptron learning algorithm (not actual perceptrons) to predict reuse of cache bkocks using features like addresses of recent memory instructions and portions of address of current block.
Verma \cite{vermaperceptron} extended the work of Teran et al. \cite{teran2016perceptron} and proposed CARP (coherence-aware reuse prediction).
CARP uses cache coherence information as an additional feature in the alogrithm of Teran et al. \cite{teran2016perceptron}.
Seng and Hamerly \cite{seng2004exploring} did one of the earliest works to present perceptron based register value predictor, where each perceptron is fed with the global history of recently committed instructions.
Later, Black and Franklin \cite{black2005neural} proposed perceptron based confidence estimator for a value predictor. 
Nemirovsky et al. \cite{nemirovsky2017machine} proposed a neural network based scheduler design for heterogeneous processors.

These algorithms have also been used to build performance/power models.
For example, Khan et al. \cite{khan2007using} proposed feed-forward neural network based predictive modeling technique to do design space exploration for chip multiprocessors.
Wu et al. \cite{wu2015gpgpu} proposed a neural network based GPU performance and power prediction model to solve the problem of slow simulation speed of simulators to study performance/power of GPUs.
The proposed model estimates performance and power consumption of applications with changes in GPU configurations.

Dai et al. \cite{dai2016block2vec} exploited deep learning techniques to propose \textit{Block2Vec} (which is inspired by Word2Vec \cite{mikolov2013distributed} used in word embeding), which can find out correlations among blocks in storage systems. Such information can be used to predict the next block accesses and used to make prefetching decisions.
Khakhaeng and Chantrapornchai \cite{khakhaeng2016finding} used perceptron based neural network to build model to predict ideal cache block size. Neural network is trained using features from address traces of benchmarks. The particular features used for training include: cache misses and size and frequency adjoining addresses which reflects temporal and spatial locality of the program.

Neural networks have also been used (both offline and online) to discover the optimal schedule of workloads or manage other aspects of shared hardware resources.
For example, Bitirgen et al.. \cite{bitirgen2008coordinated} implemented Artificial Neural Network (ANN) in the hardware for the management of shared resources in multiprocessor systems.
Each ANN (edge weights are multiplexed at the run time to achieve virtual ANNs using one hardware ANN) acts as a performance model and takes available resources and other factors describing the program behavior (e.g. read/write hits and misses in data cache and the portion of dirty cache ways allocated to the applications) as inputs and outputs the information which helps to decide which resource distiribution would result in the best overall performance.
Li et al. \cite{li2009machine} used ANNs to predict the performance of parallel tasks at run-time to do an optimal scheduling.
Nemirovsky et al.~\cite{nemirovsky2017machine} proposed an ANN based approach to perform scheduling on heterogeneous processors, which increases throughput by 25-31\% compared to a \textit{RoundRobin} scheduler on an ARM big.Little system \cite{greenhalgh2011big}.
The proposed methodology relies on ANNs to predict thread's performance for a particular scheduling interval on different hardware core types and the scheduler picks the schedule of threads that would maximize the system performance.

Other examples of the use of neural networks include the work of Ipek et al.~\cite{ipek2006efficiently} for design space exloration and the work of Chiappetta et al.~\cite{chiappetta2016real} to detect cache based side channel attacks (e.g. Flush+Reload~\cite{yarom2014flush+}) relying on the hardware performance counter values.

\subsection{The Evolutionaries:}
This class of algorithms are based on the evolutionary process of nature and rely on learning structures instead of learning parameters.
These algorithms keep on evolving and adapting to unspecified surroundings.
The most obvious example are genetic algorithms, which potentially can be used for searching the best design in a large design space.
Another problem these algorithms are a natural fit to solve is of an attacker which evolves to different versions once a preventive technique is deployed (e.g. different ML based side channel attack detection and mitigation solutions).
These algorithms can help in building preventive techniques which can evolve with the evolution of the attacker.
General purpose computer systems might be used for a diverse set of applications after being deployed.
It seems reasonable to make these systems able to evolve their configuration at runtime depending on their use.
Genetic algorithms seem to be the most suitable algorithms to learn how to adapt to the best possible system configuration depending on the workloads being executed.

There exist only a limited number of examples of the utilization of these algorithms by computer architects.
For example, ~Joel et al. \cite{emer1997language} used genetic algorithms to design better branch predictors and
Jimenez et al. \cite{jimenez2013insertion} took help of genetic algorithms to introduce a pseudo-LRU (least recently used) insertion and promotion mechanism for cache blocks in last level caches, which could result in ~5\% speedup compared to traditional LRU (least recently used) algorithm using much less overhead.
The performance of the proposed mechanism matched other contemporary techniques like DRRIP (dynamic rereference interval prediction \cite{jaleel2010high}) and PDP (protecting distance policy \cite{duong2012improving}) with less overhead.
Mukundan and Martinez \cite{mukundan2012morse} used genetic algorithms to propose MORSE (Multi-objective Reconfigurable Self-optimizing Memory Scheduler) extending Ipek et al's work \cite{ipek2008self} which used reinforcement learning for memory scheduling.
MORSE can target optimization of different metrics like performance, energy and throughput.

\subsection{The Bayesians:}
The algorithms belonging to the Bayesians are usually concerned with incorporating new evidence into previous beliefs.
The main problem they try to solve is of reducing uncertainty.
They are specially good at predicting events when there is a known probability of occurrence of particular events.
Naive Bayes and Linear Discriminant Analysis (LDA) are a couple of examples of this group of algorithms.
These algorithms have a high potential to be used by the architects due to their inherent characterstics, as a lot of architecture problem solutions rely on the knowledge of existence of certain events.
For instance, a bayesian based model to detect hardware side channel attacks can rely on the fact that the probability of system under no attack would be very high normally.
Similarly, a model to predict memory events can depend on the average probability of an instruction being a memory operation.
Beckman and Sanchez \cite{beckmann2017maximizing} recognized that the problem of design of cache replacement policies have to deal with uncertainty as the time when the candidate cache blocks will be accessed is not known.
Thus, bayesians have the potential to learn in such environments and develop smart cache replacement policies.

Overall, Bayesians also have not found a lot of use in the computer architecture community as our survey suggests.
Jung and Pedram \cite{jung2010supervised} used Bayesian classification to tune voltage and frequency settings to reduce system's energy consumption in a multicore processor.
Reagen et al. \cite{reagen2017case} used Bayesian optimization \cite{pelikan1999boa} to explore design space for a Deep Neural Network hardware accelerator.
Different design parameters studied in this work, like neurons/layer for DNN, L2 regularization for DNN, learning rate of DNN, loop parallelism in hardware, and hardware pipelining, have "complex interactions" among them which the bayesian optimization tries to learn.

\subsection{The Analogizers:}
Analogizers believe in learning things from other examples.
The primary problem they target is finding similarities between different situations.
Few examples of algorithms belonging to this domain include: support vector machines (SVM), k-means clustering and k-nearest neighbor (KNN).
Some of these algorithms (like k-means clustering and KNN) are used in un-supervised learning.
Algorithms from this group can be good candidates to be applied in cutting down the search space for design space exploration especially when configuration space is large and there exists multiple
designs with marginal differences.
These algorithms can help to focus on designs which are sufficiently different from a baseline.
Similarly, they can help in detecting similarities at the granularity of programs or at the granularity of instructions to guide better predictions at the run-time by different microarchitectural structures.

Analogizers have been pretty successful in the domain of computer architecture.
For example, Culpepper and Gondree \cite{culpepper2005svms} used support vector machines (SVM) to improve accuracy of branch prediction.
This SVM based branch predictor performed better compared to fast-path based predictor and gshare predictor at high hardware budgets \cite{culpepper2005svms}.
Sherwood et al. \cite{sherwood2002automatically} used K-means clustering to form groups/clusters of similar basic block vectors (BBV~\footnote{BBV contains the frequency of occurrence of all basic blocks during an interval of execution}).
These groups then act as a few representative portions of the entire program and are known as \textit{Simpoints}, which can be used to approximate the performance/power for the entire program.
Hoste et al. \cite{hoste2006performance} detected similarity among programs using data from various microarchitecture-independent statistics (like instruction classes, instruction-level parallelism and register traffic, branch predictability and working set sizes) to predict performance of programs similar to reference benchmarks with known performance on specific microarchitecture.
They used principal component analysis for this purpose alongwith other ML techniques.

Walker et al. \cite{walker2017accurate} relied on heirarchical cluster analysis \cite{bridges1966hierarchical} to form groups of various performance monitoring counters which are then used to estimate power consumption for mobile and embedded devices and showed that their technique leads to lower percentage error compared to many other power estimation techniques~\cite{pricopi2013power,walker2015run,rethinagiri2014system,rodrigues2013study}.

Baldini et al. \cite{baldini2014predicting} trained two binary classifiers Nearest Neighbor with Generalized exemplars (NNGE) and Support Vector Machine (SVMs) to predict
possible GPU speed-up given a run of parallel code on CPU (using OpenMP implementation).
Wang and Ipek \cite{wang2016reducing} proposed an online data clustering based technique to reduce energy of data transfers in memory by reducing the number of ones in the data.
\subsection{Reinforcement Learning: The Sixth Tribe~\protect\footnote{We refer to reinforcement learning as the sixth tribe of ML as it is one of the most important algorithms belonging to the class of self learning algorithms. Reinforcement learning does not belong to the original ML tribes taxonomy of Pedro Domingos~\cite{domingos2015master}.}}

Reinforcement learning refers to the process in which an agent learns to take actions in a given environment through independent exploration of different possible actions and choosing the ones that increase the overall reward.
Reinforcement learning does not require any prior data to learn to take the right action, rather it learns to do this on the fly.
This makes it a good fit to be used for computer architecture problems where significant former knowledge or data is not available and the right action can only be learned dynamically.
For example, hardware optimizations or hardware based resource management techniques which are totally transparent to the software can rely on reinforcement learning to decide the best action to take for the overall efficiency of the system.

One of the earliest examples of the use of reinforcement learning in computer architecture is the work of Ipek et al \cite{ipek2008self}.
Ipek et al \cite{ipek2008self} introduced reinforcement learning based DRAM scheduling for an increased utilization of memory bandwidth.
The DRAM scheduler which acts as a reinforcement learning agent utilizes system state defined by different factors like number of read/write requests residing in the transaction queue. The actions that the agent can take include all normal commands of DRAM controller like read, write, pre-charge and activate.
Ipek et al \cite{ipek2008self} implemented a five stage hardware pipeline to calculate q-values~\cite{sutton1998reinforcement} associated with the credit assignment which determine the eventual benefit for an action given the current state.

Peled et al \cite{peled2015semantic} used reinforcement learning to approximate program semantic locality, which was later used to anticipate data access patterns to improve prefetching decisions.
Accesses are considered to have semantic locality if there exists a relationship between them via a series of actions.
A subset of different attributes (e.g. program counter, accesses history, branch history, status of registers, types and offsets of objects in program data structures and types of reference operations) is used to represent the current context a program.

Juan and Marculescu~\cite{juan2012power} proposed a reinforcement learning based DVFS technique which divides the central agent in many "distributed" agents and uses a supervisor for coordination among other agents to maximize performance under given power budget.
Similarly, Ye and Xu \cite{ye2014learning} proposed a reinforcement learning based dynamic power management (DPM) technique for multi-core processors.


\subsection{Regression Techniques -- Statistics Meets Machine Learning~\protect\footnote{Regression techniques are not a part of the original ML tribes taxonomy of Pedro Domingos~\cite{domingos2015master}}}

Regression techniques from applied statistics have been largely borrowed by machine learning.
A detailed discussion on different regression techniques used in machine learning is provided in the Appendix section.
In this section, we take a look at how these regression techniques have influenced the field of computer architecture.

Performance and power estimation of different workloads on a particular architecture is a
critical step to design new architectures or to have a better understanding of the already existing
architectures.
There are numerous examples of research works where regression techniques are used
to estimate performance or power consumption.
Performance/power estimation of applications on a particular architecture is mostly done by simulation tools and analytical models.
Regression techniques have also been used to build new and accurate models using empirical data.
Sometimes they are also used to increase the accuracy of simulation techniques.
For example, Lee et al. \cite{lee2015powertrain} proposed regression analysis based calibration methodology called \textit{PowerTrain} to improve the accuracy of McPAT~\cite{li2009mcpat}.
McPAT~\cite{li2009mcpat} is a well-known power estimation simulator, but it is shown to have various inaccuracies due to different sources like un-modeled architectural components, discrepancy between the modeled and actual hardware components and vague configurational parameters \cite{xi2015quantifying}.
The proposed calibration methodology called \textit{PowerTrain} uses power measurements on real hardware to train McPAT using regression analysis.
Reddy et al. \cite{reddy2017empirical} studied correlation among gem5~\cite{binkert2011gem5,lowe2020gem5} statistics and hardware performance monitoring counters to build a model in gem5 to estimate power consumption of ARM Cortex-A15 processor. This work uses the same model built by Walker et al. \cite{walker2017accurate} but does not include all performance monitoring events as some gem5 equivalents would not be available. The used events in ML model are cycle counts, speculative instructions, L2 cache accesses, L1 instruction cache accesses and memory bus reads. Reddy et al. \cite{reddy2017empirical} show that the differences between statistics of the simulator (gem5) and those of the real hardware only affect the estimated power consumption by approximately 10\%.

There are also examples \cite{lee2008cpr,pricopi2013power} of use of regression techniques to estimate performance of multi-core processors from single core processors.
Lee et al. \cite{lee2008cpr} used spline-based regression to build model for multiprocessor performance estimation from uniprocessor models to mitigate the problems associated with cycle accurate simulation of multiprocessors.
Pricopi et al.~\cite{pricopi2013power} used regression analysis to propose an analytical model to estimate performance and power for heterogeneous multi-core processors (HMPs). During an application run, a cpi (cycles per instruction) stack is built for the application using different micro-architectural events that can impact the execution time of the application. Relying on some compiler analysis results along-with the cpi stack, performance and power can be estimated for other cores in an HMP system. Experiments performed with an ARM big.Little system indicate an intra-core prediction error of below 15\% and an inter-core prediction error of below 17\%.

Regression techniques are also used in heterogeneous systems for cross platform performance estimation. Examples include ~\cite{zheng2015learning}, \cite{boran2016performance}, \cite{o2017gpu}, and \cite{ardalani2015cross}.
Zheng et al. \cite{zheng2015learning} used Lasso Linear Regression and Constrained Locally Sparse Linear Regression (CLSLR) to explore correlation among performance of same programs on different platforms to perform cross-platform performance prediction. The test case used in their work is prediction of an ARM ISA platform based on performance on Intel and AMD x86 real host systems.
 The authors extended this work in \cite{zheng2016accurate} by proposing \textit{LACross} framework which applies similar methodology to predict performance and power at fine granularity of phases. \textit{LACross} is shown to have an average error less than 2\% in entire program's performance estimation, in contrast to more than 5\% error in \cite{zheng2016accurate} for SD-VBS benchmark suite \cite{venkata2009sd}.
Boran et al. \cite{boran2016performance} followed Pricopi et al.'s \cite{pricopi2013power} work and used regression techniques to estimate execution cycle count of a particular ISA core based on the performance statistics of another ISA core. This model is used to dynamically schedule programs in a heterogeneous-ISA multi-core system. ARMv8 and x86-64 based multi-core system is used to validate the model. Although this model shows above 98\% accuracy to estimate performance on any particular ISA core, the inter-core performance estimation has high error (23\% for estimation from ARM to x86 and 54\% for estimation from x86 to ARM).
O'Neal et al. \cite{o2017gpu} used various linear/non-linear regression algorithms to propose GPU performance predicting model focusing on "pre-silicon design" of GPUs using DirectX 3D workloads.
Ardalani et al. \cite{ardalani2015cross} proposed \textit{XAPP} (\textit{Cross Architecture Performance Prediction}) technique to estimate GPU performance from CPU code using regression and bootstrap aggregating (discussed in the Appendix section). Different features associated with the program behavior are used to train machine learning models. These features include some basic features related to ILP (instruction level parallelism), floating point operations and memory operations and also some advanced features like shared memory bank utilization, memory coalescing and branch divergence.

Regression techniques have also been used in design space exploration (e.g. ~\cite{lee2007illustrative, jia2012stargazer}).
Lee et al. \cite{lee2007illustrative} used regression modeling for fast design space exploration to avoid expensive cycle accurate simulations.
Jia et al. \cite{jia2012stargazer} proposed \textit{StarGazer}, an automated framework which uses a stepwise regression algorithm to explore GPU design space by using only few
samples out of all possible design points and estimates the performance of design points with only 1.1\% error on average.

Some regression techniques have also been used in the domain of hardware security, specifically to detect the malware or micro-architectural side channel attacks.
For instance, Ozosoy et al. \cite{ozsoy2016hardware} used neural network and logistic regression for detection of malware using hardware architectural features (like memory accesses and instruction mixes).
They evaluated the FPGA based implementations of both ML models (logistic regression and neural networks).
Khasawneh et al. \cite{khasawneh2017rhmd} used neural networks and logistic regression (LR) based hardware malware detectors to prove the possibility of evading malware detection. Availability of malware detector training data makes it possible for attackers
to reverse engineer detectors and potentially modify malware to evade their detection.
Khasawneh et al. \cite{khasawneh2017rhmd} also proposed randomization
based technique to attain resilient malware detection and avoid reverse engineering.
Many other works which utilize regression or other ML techniques for micro-architectural side channel attack detection are referred in~\cite{akram2020meet}.

\section{Summary Table}

Table~\ref{tab:summary} provides a summary of the previously discussed literature survey at a finer granularity of individual ML techniques.
Each column in this table refer to a broader category (or field) of computer architecture that has relied on ML.

\begin{small}
\begin{center}
\begin{landscape}
 \begin{longtable}{| m{2.3cm} | m{2.5cm} | m{2.5cm} | m{2.1cm} | m{2cm} | m{2cm} | m{1.8cm} | m{1.8cm} |}
 \caption{Summary of the Literature Survey on Machine Learning in Computer Architecture}
  \\ \hline
  \rowcolor{blue!25}\textbf{Technique} & \textbf{Microarchitecture} & \textbf{Power/Perform. Estimation} & \textbf{Thread\newline Scheduling} & \textbf{Design Space Exploration} & \textbf{Energy\newline Improvement} & \textbf{Instruction Scheduling} & \textbf{Hardware Security} \\ \hline
  \endfirsthead
    \multicolumn{8}{|c|}{{\textbf{Note:} The examples of sub-problems for each column in this table are following: \textbf{Microarchitecture}: branch prediction, cache replacement,}} \\
\multicolumn{8}{|c|}{{cache reuse prediction, value prediction, memory scheduling, prefetching, network-on-chip design. \textbf{Power/Perform. Estimation:} power/performance}} \\
\multicolumn{8}{|c|}{{estimation for single or multi-cores, gpus, cross-platform (e.g. one ISA to other or cpu to gpu), simulation augmentation. \textbf{Thread Scheduling:}}} \\
\multicolumn{8}{|c|}{{management of shared hardware resources, thread scheduling in heterogeneous systems. \textbf{Design Space Exploration:} design space exploration of}} \\
\multicolumn{8}{|c|}{{single or multi-cores, gpus, accelerators, network-on-chips. \textbf{Energy Improvements:} energy aware thread assignment or micro-architecture design,}} \\
\multicolumn{8}{|c|}{{ dynamic voltage and frequency scaling. \textbf{Instruction Scheduling:} static/dynamic instruction scheduling. \textbf{Hardware Security:} micro-architectural}} \\
\multicolumn{8}{|c|}{{side-channel attack detection and evasion.}} \\ \hline
    \endfoot
    \textbf{Decision Trees} & \cite{fern2000dynamic,liao2009machine,rahman2015maximizing,navarro2017machine}  & \cite{carvalho2020using,elmoustapha2007comparison,jundt2015compute}  & \cite{helmy2015machine,prodromou2019deciphering} & \cite{moeng2010applying,sen2019machine,zhu2020bbs} & & \cite{monsifrot2002machine,malik2008learning} & \cite{mushtaq2020whisper,mushtaq2018nights} \\ \hline
    \textbf{Genetic Algo.} & \cite{emer1997language}  & \cite{hoste2006performance} & & \cite{stanley1995parallel,reagen2017case} & & \cite{cooper1999optimizing,leather2009automatic} & \\ \hline
    \textbf{K-Nearest\newline Neighbor}  & \cite{liao2009machine} & \cite{carvalho2020using,baldini2014predicting} & \cite{helmy2015machine} & & & & \\ \hline
    \textbf{K-Means}  & & \cite{sherwood2002automatically,jundt2015compute,wu2015gpgpu} &  & & \cite{ma2013edoctor,wang2016reducing} & & \\ \hline
    \textbf{Linear/Non-Linear\newline Regression}  & & \cite{agarwal2019performance,o2018hardware,o2018hlspredict,elmoustapha2007comparison,pricopi2013power,zheng2015learning,zheng2016accurate,lee2015powertrain,ardalani2015cross,boran2016performance,o2017gpu,walker2017accurate,reddy2017empirical} & \cite{mishra2017esp,prodromou2019deciphering} & \cite{lee2007illustrative,jia2012stargazer} & \cite{yang2015adaptive} & & \cite{mushtaq2020whisper,mushtaq2018nights} \\ \hline
    \textbf{Logistic\newline Regression} & \cite{liao2009machine,rahman2015maximizing} & \cite{carvalho2020using} & & \cite{cochran2011pack,lopes2020machine} & & & \cite{ozsoy2016hardware,patel2017analyzing,khasawneh2015ensemble,khasawneh2018ensemblehmd,khasawneh2017rhmd,mushtaq2020whisper}, \\ \hline
    \textbf{Bayes Theory} & \cite{navarro2017machine} & &  & \cite{khan2007using,reagen2017case} & \cite{jung2010supervised} & & \cite{patel2017analyzing} \\ \hline 
    \textbf{Neural Network} & \cite{zhang2020dynamic,yin2020experiences,doudali2019kleio,tarsa2019improving,shineural,linneural,garza2019bit,bhatia2019perceptron,tarsa2019post,shi2019applying,peled2019neural,bhardwaj2018pecc,calder1997evidence,jimenez2002neural,wang2005dynamic,jimenez2003fast,seznec2014tage,mao2017study,teran2016perceptron,dai2016block2vec,khakhaeng2016finding,yazdanbakhsh2015neural,margaritov2018virtual} & \cite{renda2020difftune,foots2020cpu,carvalho2020using,mendis2019ithemal,o2018hlspredict,elmoustapha2007comparison,wu2015gpgpu,jhacache} & \cite{bitirgen2008coordinated,li2009machine,wang2009mapping,wang2014integrating,helmy2015machine,nemirovsky2017machine,gomatheeshwari2020appropriate} & \cite{ipek2006efficiently,khan2007using,braun2019understanding,lin2019design,li2019characterizing,sen2019machine,enef,lopes2020machine} & \cite{won2014up} & \cite{jainlearning} & \cite{ozsoy2016hardware,patel2017analyzing,khasawneh2018ensemblehmd,khasawneh2017rhmd,chiappetta2016real,mirbagher2020perspectron,mushtaq2020whisper} \\ \hline

   \textbf{PCA} & & \cite{hoste2006performance} & & & & & \\ \hline
 \textbf{Random\newline Forest} & & \cite{ardalani2019static,o2018hardware,o2018hlspredict,mariani2017predicting}  & \cite{prodromou2019deciphering} & \cite{sen2019machine} & & & \\ \hline
    \textbf{Reinforcment Learning} & \cite{wang2020cure,kim2020autoscale,zheng2019energy,sanchez2019cads,ipek2008self,peled2015semantic} & & \cite{fettes2020hardware} & & \cite{juan2012power,ye2014learning,fettes2018dynamic}  & \cite{mcgovern1999scheduling} & \\ \hline
    \textbf{SVM} & \cite{culpepper2005svms,liao2009machine} & \cite{elmoustapha2007comparison,baldini2014predicting} & \cite{wang2009mapping,wang2014integrating,helmy2015machine} & \cite{sen2019machine} & & & \cite{patel2017analyzing,mushtaq2020whisper,mushtaq2018nights} \\ \hline
    \textbf{Others} & \cite{vietri2018driving,jimenez2005piecewise,navarro2017machine} & \cite{elmoustapha2007comparison,walker2017accurate} & \cite{mishra2017esp} & \cite{reagen2017case} & \cite{li2018processor} & \cite{agakov2006using} & \cite{patel2017analyzing} \\ \hline
 \end{longtable}
\label{tab:summary}
\end{landscape}
\end{center}
\end{small}
\clearpage

\section{Opportunities and Challenges:}

This section presents some opportunities to use ML for architecture that have not been explored in their full potential if explored at all.
Moreover, we also enlist a number of challenges that need to be tackled to be able to fully exploit ML for architecture research.

\subsection{Opportunities:}
\begin{itemize}

\item There exists an opportunity to borrow ideas from ML instead of using ML algorithms in their intact form.
Since, implementing ML algorithms in the hardware can be costly, computer architects can rely on ML algorithms to optimize the classical learning techniques used in computer systems.
For instance, the work done by Teran et al. \cite{teran2016perceptron} to apply perceptron learning algorithm to predict reuse of cache blocks is a good example of this.
\item ML based system components can be used in synergy with traditional components. A good example of this collaboration is the work of Wang and Luo \cite{wang2017data} to perform data cache prefetching discussed in the previous section.
\item ML techniques can potentially be used to build customizable computer systems which will learn to change their behavior at different granularities.
Take an example of an 'A' cache eviction policy which works better for workload type 'X' and another policy 'B' which works better for workload type 'Y'.
Since, ML makes learning easier, same processor can learn to adapt to using policy 'A' or 'B' depending on the workloads run on it.
\item Computer architects have relied on heuristics to design different policies.
One possiblity is to start using ML in place of these heuristics.
Machine Learning can also help us to come up with new and better heuristics that are hard to ascertain otherwise.
\item For all the decisions taken inside a computer system, multiple policies are used.
If single (or a small number of) ML algorithm(s) can learn all of these policies, the design of systems can be significantly simplified.
\item As we observed earlier, performance and power estimation of existing or new architectures is an area which can be highly influenced by ML.
Architects have historically used simulation tools or mathematical models for this purpose.
Future architectural innovations are expected to come from optimizations across the entire
software/hardware computing stack~\cite{leiserson2020there}.
Current simulation techniques are not designed with this kind of use case in mind.
Therefore, ML techniques can come to rescue and enable building simulation tools of the future as traditional cycle level modeling might be too slow for cross-stack studies.
Not only that ML techniques can enable building new tools, they can help in improving the accuracy of current simulation tools (or performance/power models) which are the primary enabler of computer architecture research at the moment (for example the work by Renda et al.~\cite{renda2020difftune}).
\item ML can specially prove useful to build systems/accelerators for ML.
\item Design space explorations (which are extensively done in computer architecture research) should use more of ML techniques (especially the ones belonging to evolutionaries and analogizers).
Techniques like genetic algorithms can help to evolve the design space itself as well, thus enabling the evaluation of significant configurations which might remain hidden otherwise.
Since, all of this exploration is done statically (is part of pre-silicon design), there will be no issues of the cost of the algorithm here.
\item Computer architecture research often relies on combination of multiple (independently functioning) structures  to optimize the system. For example, tournament branch predictors use multiple branch predictors and pick one of them for a given prediction.
Similar tasks can be relegated to ML to learn the best of different independent techniques.
\item Computer architects often run a number of benchmarks/workloads on a multitude of hardware configurations (and generate multiple statistics) when studying new or existing ideas.
This leads to huge amounts of data which researchers often study in a limited fashion on ad hoc basis.
In contrast, if ML based analysis is performed on this data, we might be able to see the true potential of that data and its ability to help us discover new findings.
\end{itemize}


\subsection{Challenges:}
\begin{itemize}
\item The resources needed by these ML techniques when they are implemented in hardware can be significant.
Specially, if these techniques need to be used for microarchitectural decisions at a finer granularity (at the program phase level), the hardware implementations become more critical.
There is a need to develop more optimized hardware implementations of these algorithms to make them more feasible to be applied towards architectural problems/design.
\item ML algorithms primarily rely on huge data sets for learning.
In comparison to other fields, there does not exist standardized data sets in architecture domain to be used by ML algorithms.
\item Although there exist a number of research works which take help of ML algorithms for computer architecture problems, the use of a particular algorithm is not justified by the authors mostly.
As we observed in this paper, there are ML algorithms which naturally fit certain problems.
We think it will be sensible to emphasize the use of appropriate algorithms for a given problem, which can help in ensuring the robustness of the proposed ideas.
\item ML methods have found more use in branch prediction like problems where the decision to be
made is a binary decision (taken/not-taken) as opposed to the problems like memory prefetching
where the output has to be a complete memory address (which increases the complexity of the ML
model).
\item Data selection, cleaning, and pre-processing and then interpreting results is a challenge.
\item Formulation of architecture problems as machine learning problem is a general challenge.
\item As we pointed out in the opportunities sub-section that ML can help improving the accuracy of simulation tools, that opportunity comes with a challenge of building new tools
to embed ML algorithms into simulation techniques. Specifically, the question of how to make the simulation tools talk to ML algorithms needs to be addressed.
\item There exist a big disparity between the time that ML algorithms take to make a prediction and the time that often microarchitectural structures take for an estimation (known as time scale gap). This needs to be solved before practical ML based microarchitectural solutions can be designed.
An example where this problem has been addressed is the design of branch predictors i.e. the proposal of analog versions of neural network based branch predictors (for example: \cite{jimenez2011optimized,amant2009mixed}) to improve latency of prediction.

\item It is also not clear how (and if) the ML based architecture design will scale.
Traditionally, in computer architecture, many microarchitectural/architectural techniques work the same way irrespective of the workloads used or size of the processor.
For heuristics based designs, it is easier for humans to argue about their scalability.
However, this might not be true for ML based design (specially if it is not very interpretable).
Thus, the question arises if the optimal machine learning based design for a specific problem at hand will work if the workloads change significantly or the hardware resources scale to a bigger budget.
\item As ML is finding more and more uses in the real world, many security challenges have been raised.
      Before computer architecture design can be fully influenced by ML, the attacks (like adversarial ML attacks) possible on ML techniques need to be addressed
      in the context of hardware design.
      Moreover, these machine learning implementations might lead to new side-channels and computer architects will need to be cognizant of this to avoid major security vulnerabilities in the future.
\end{itemize}

\section{CONCLUSION}
This paper surveys a wide set of representative work in computer architecture which utilize ML techniques.
We identify the fundamental properties of different classes of Machine Learning and their relation with the problems in the domain of computer architecture.
It is observed that the usage of ML techniques is on the rise and future computer architecture research can further leverage the exciting developments from the field of Machine Learning. However, a number of challenges need to be addressed to fully exploit the ML potential for computer architecture research.

\bibliographystyle{acm}
\bibliography{sample-bibliography}

\appendix
\section{An Overview of ML Techniques}
\label{sec:overview}

Machine Learning (ML) refers to the process in which computers learn to make decisions based on the given data set without being explicitly programmed to do so \cite{alpaydin2014introduction}.
There are numerous ML algorithms that can be grouped into three categories: supervised learning algorithms, unsupervised learning algorithms and other types of algorithms. Learning algorithms are usually trained on a set of samples called a "training set". A different set of data is usually used to test the accuracy of decisions made by the learning algorithm known as "test set". 

\begin{figure}[htbp]
\centering
\includegraphics[scale=0.5]{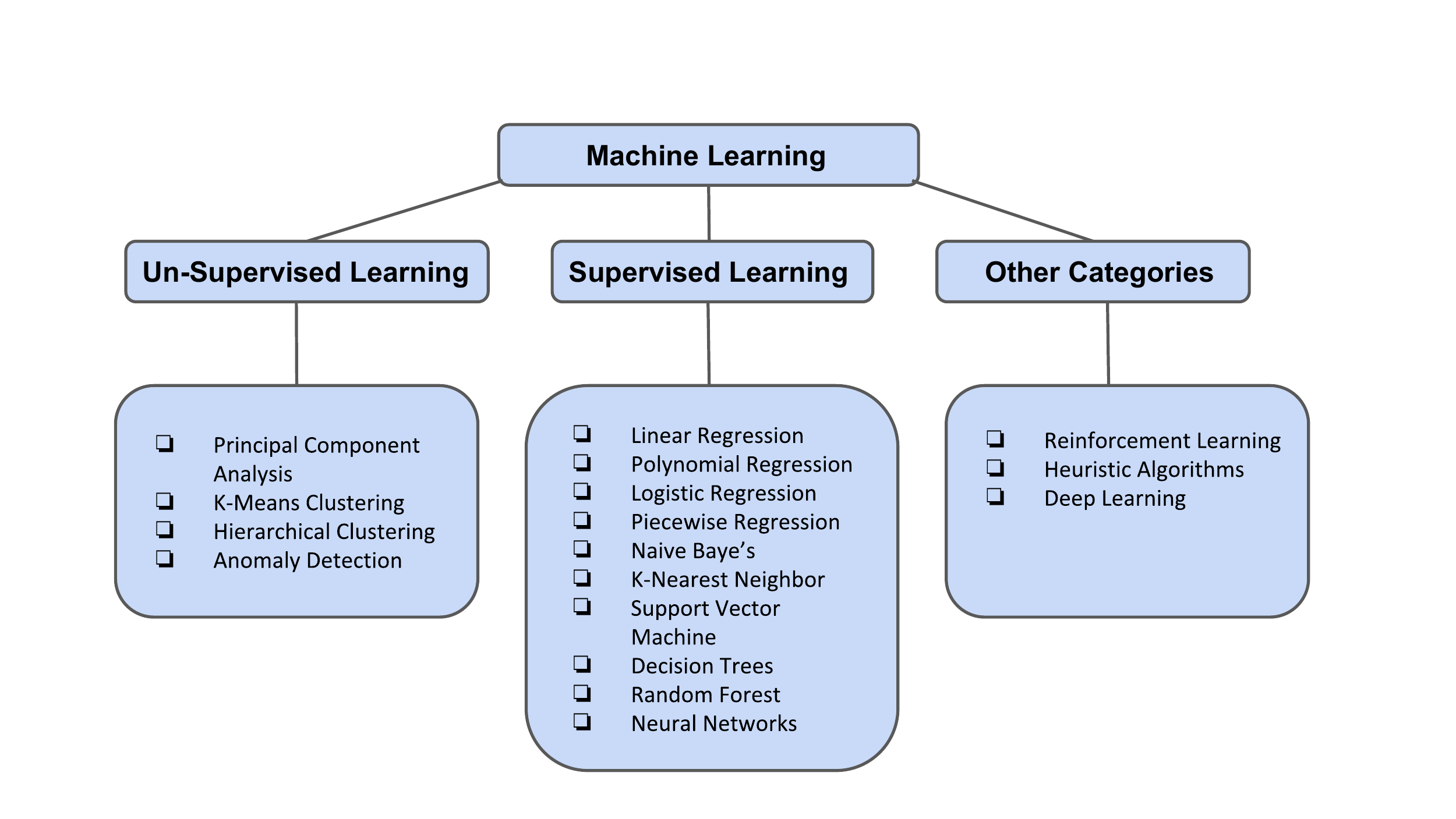}
\caption{Classification of common ML algorithms}
\label{fig:Overview}
\end{figure}

This section briefly discusses different ML techniques that are used by the research surveyed in this paper. Readers who wish to have detailed explanation of the discussed methodologies in this section can refer to \cite{james2013introduction}. A common textbook classification of ML algorithms is shown in Figure~\ref{fig:Overview}.

\subsection{Supervised Learning}
The process of training machine learning models given a set of samples of input vectors with the corresponding output/target vectors is called Supervised learning \cite{christopher2016pattern}. The trained model can then be used to predict ouput vectors given a new set of input vectors. The output/target vectors are also known as labels and input data as labelled data.

Machine learning models can operate on different types of data. If the target labels in a data set can be a set of discrete classes/categories, the modeling task is referred as classification task. On the other hand if the target labels in a data set are continous variables, the modeling task is called a regression task \cite{james2013introduction}. In other words, in classfication the built models predict qualitative response and in regression the built models predict quantitative response. 


Many of the ML algorithms can be applied to both regression and classification tasks, while there are others that can only be used for classification or regression tasks. 

\subsubsection{Linear Regression}
Linear regression is used to relate different variables and approximates the effect of changing a variable on other variables. Assume that there exists a predictor variable X (also known as independent/input variable) and a corresponding quantitative response Y (also known as dependent/output variable). Linear regression, assuming that there exists a linear relation between dependent and independent variables, generates a model as shown in (1) \cite{james2013introduction,seber2012linear}.

\begin{equation}
Y = \alpha_0 + \alpha_1X
\end{equation}

Here $\alpha_0$ and $\alpha_1$ are regression coefficients unknown before training the model using given input data.

\subsubsection{Multiple Linear Regression}
In case the ouput variable Y is dependent on more than one independent variables or predictors, the linear regression equation can be extended to include all those predictors. Assuming there are p independent variables the multiple linear regression model would be given as in (2) \cite{james2013introduction}.

\begin{equation}
Y = \alpha_0 + \alpha_1X_1 + \alpha_2X_2 + \alpha_3X_3 + \alpha_4X_4 + .... + \alpha_mX_m
\end{equation}

Here $\alpha_0$ to $\alpha_m$ are regression coefficients unknown before training the model using given input/output data.

\subsubsection{Polynomial Regression}

If there exists a non-linear relation between response and predictor variables, polynomials of predictor variables can be added in the regression model resulting into Polynomial Regression. 
For example if a quadratic relation exists between input X and output Y the regression model would become as in (3) \cite{james2013introduction}.

\begin{equation}
Y = \alpha_0 + \alpha_1X + \alpha_2{X}^2
\end{equation}


\subsubsection{Logistic Regression}
Logistic regression algorithm finds its use purely for classification problems. In case of logistic regression the built model predicts the probability that the response Y would be member of particular class. Given an input/independent variable X, logistic regression model can be represented by the logistic function of (4) \cite{james2013introduction}.

\begin{equation}
p(X) = e^{\alpha_0+\alpha_1X}/(1+e^{\alpha_0+\alpha_1X})
\end{equation}
 
This relation can also be extended to multiple logistic regression in case of more than one independent variables as in the case of multiple linear regression. 

\subsubsection{Piecewise Polynomial Regression} 
Piecewise polynomial regression tries to avoid use of high degree polynomials. Instead it uses models built by low-degree polynomials, each corresponding to a different region of the variable X. 
As an example, "piecewise cubic polynomial" will use cubic regression models shown in (5) and (6) for different ranges of x. The points of changes in values of coefficients are known as knots. The given example in (5) and (6) has a single knot at point c.

\begin{equation}
y_i = \alpha_{01} + \alpha_{11}x_i + \alpha_{21}{x_i}^2 + \alpha_{31}{x_i}^3 + \epsilon_i  \hspace{1cm} if \hspace{2mm} x_i < c;
\end{equation}
\begin{equation}
\hspace{1mm} y_i = \alpha_{02} + \alpha_{12}x_i + \alpha_{22}{x_i}^2 + \alpha_{32}{x_i}^3 + \epsilon_i  \hspace{1cm} if \hspace{2mm} x_i >= c;
\end{equation}


\subsubsection{Naive Bayes'} 

Bayes' theorem is used significantly to compute conditional probability. Mathematical form of bayes' theorem is given in (7) \cite{vapnik1998statistical}. 

\begin{equation}
Pr(h|e) = Pr(e|h)Pr(h)/Pr(e)
\end{equation}

Here Pr(h|e) represents the probability that the hypothesis h will be true if some event e exists (posterior probability). Pr(e|h) represents the probability that the event e would occur in presence of an hypothesis h. Pr(h) is the proability of hypothesis by itself without the event (prior probability) and Pr(e) is the probability of the event e \cite{vapnik1998statistical}. Bayes' theorem is largely used for classification purposes in ML. Bayes' theorem based classifiers that assume that the given inputs/features are independent of each other are known as \textit{Naive Bayes' Classifiers}. Usually, a small number of data points are sufficient to train Naive Bayes' classifiers. They are famous for their scalability and efficiency. A more detailed review of Bayes' classifiers can be found in \cite{murphy2006naive}.


\subsubsection{K-Nearest Neighbors}

K-Nearest Neighbors (KNN) is used to classify unseen data points by observing K already-classified data points that are closest to the unseen data point. Different distance measures can caclculate the distance of new/unseen data point from neighboring data points such as Euclidean, Manhattan, Hamming etc.~A mathematical representation of KNN is given in (8). Assuming that there is a test observation that needs to be classified, and K closest points are represented by $N_0$, then to find out if test point belongs to class j KNN relies on (8). In (8), a fraction of points in $N_0$ belonging to class j gives the conditional probability of class j.

\begin{equation}
1/K\sum_{i \epsilon N_0} I(y_i=j)
\end{equation}


\begin{figure*}[htbp]
\begin{minipage}{.5\textwidth}
  \centering
\centerline{\includegraphics[scale=0.6]{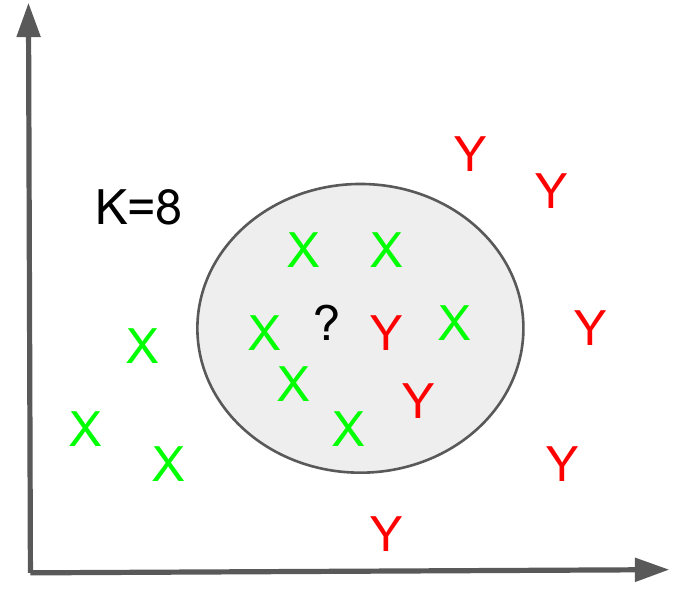}}
\caption{KNN}
\label{fig:KNN} 
\end{minipage}%
\begin{minipage}{.5\textwidth}
  \centering
\centerline{\includegraphics[scale=0.6]{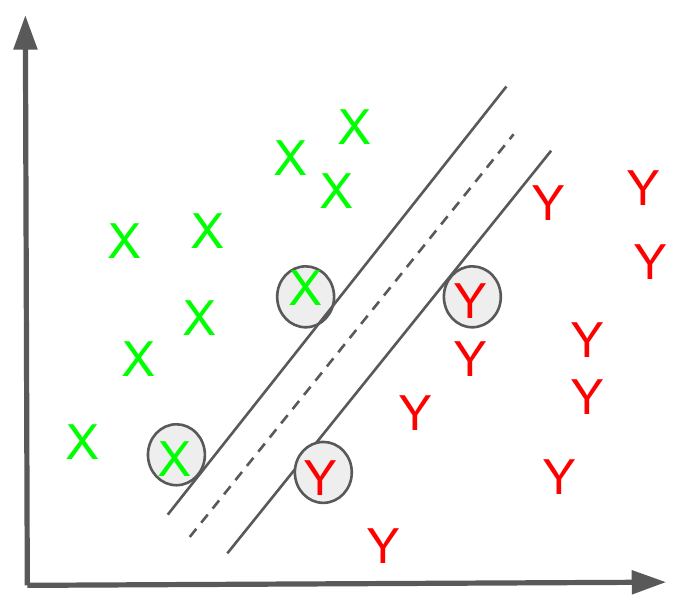}}
\caption{SVM}
\label{fig:SVM}
\end{minipage}
\end{figure*}

Figure \ref{fig:KNN} shows an example of how KNN algorithm works. For K=8, the data point in question will be classified as `X' as most of its neighbors belong to category `X'. ~KNN is specifically useful for ML problems where "underlying joint distribution of the result and observation" is unknown. KNN does not scale well with larger data sets since it makes use of the complete training data to make new predictions for test data. KNN has also been used for regression problems \cite{christopher2016pattern}. More information on KNN can be found in \cite{cover1967nearest}.

\subsubsection{Support Vector Machines}
Support Vector  Machine (SVM) is a non-probabilistic ML model that has been used for both classification and regression problems, but mostly for classification. Assuming that we have n number of features or input variables, SVM classifier plots given data points in an n-dimensional space. The algorithm finds hyper-planes in this n-dimensional space which can distinguish given classes with the largest margin. This hyperplane can be a linear/non-linear function of input variables resulting into linear/non-linear SVM. SVM based classifiers use "a few training points" (also known as support vectors) while classifying new data points. Theoretically, support vectors are the most difficult points to classify since they are the nearest to the "decision surface". Maximum margin between these support vectors ensures that the distance between classes is maximized. Figure \ref{fig:SVM} shows a linear SVM classifier, distinguishing between two classes of data.

SVMs are effective for high number of features and are memory-efficient. However, these models are not very interpretable \cite{kim2003constructing}. More information on SVMs can be found in \cite{james2013introduction,boser1992training}

\subsubsection{Decision Trees} 
\begin{figure*}[htbp]
\begin{minipage}{.5\textwidth}
  \centering
\centerline{\includegraphics[scale=0.5]{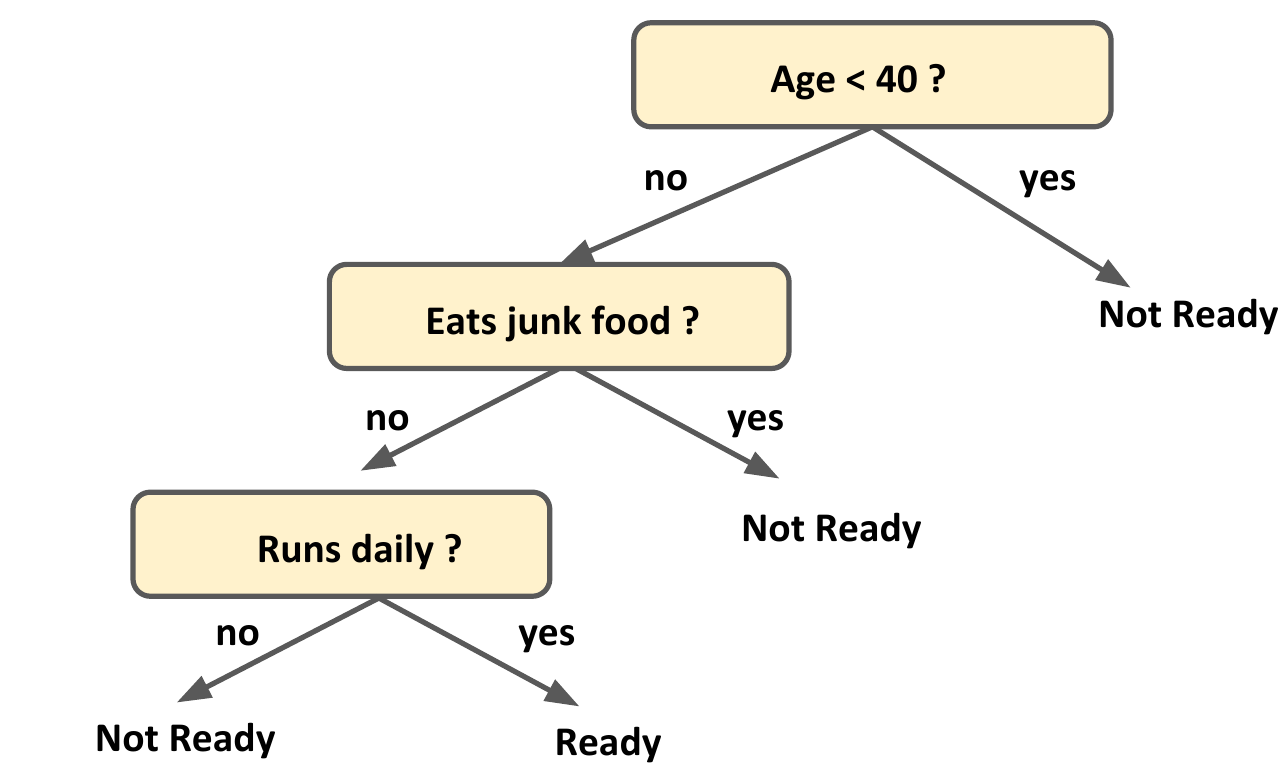}}
\caption{Decision Tree}
\label{fig:DTree}
\end{minipage}%
\begin{minipage}{.5\textwidth}
  \centering
\centerline{\includegraphics[scale=0.6]{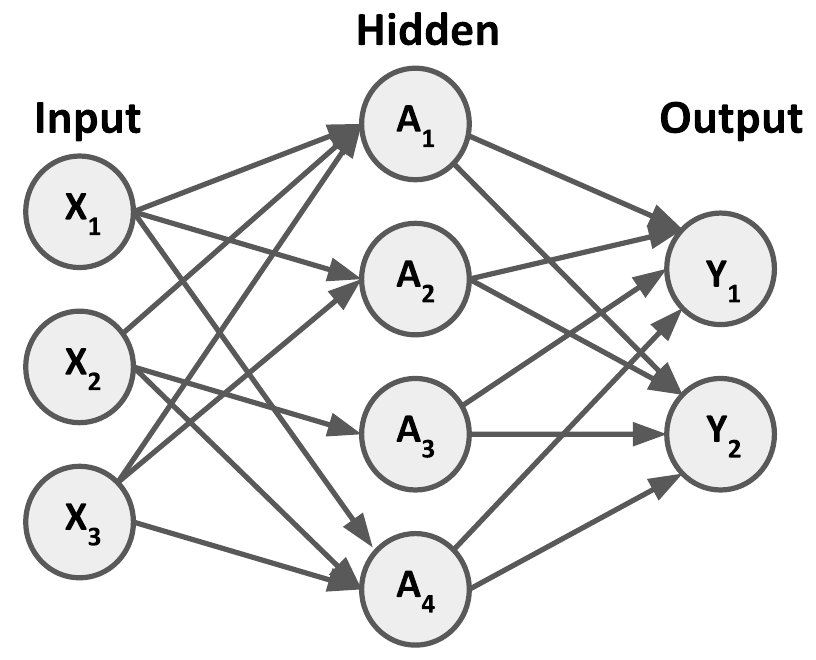}}
\caption{Neural Network}
\label{fig:NNet}
\end{minipage}
\end{figure*}

Decision trees summarize the relationship between different values of given features (represented by branches) and the conlusions/results about the output variables's value (represnted by leaves) \cite{DTree}. Decision trees can be used for both classification and regression problems. Figure \ref{fig:DTree} shows an example of a decision tree used for a classification problem; is a person ready to run a marathon?

Decision trees are easy to interpret because of their pictorial nature. Generally decision trees are considered to be less accuarate compared to other advanced ML techniques. Interested readers can obtain more information about decision trees in \cite{james2013introduction,breiman2017classification}.

\subsubsection{Random Forests} 
Random forests form a large number of decision trees based on the training data and then uses the combined behavior of those trees to make any predictions \cite{RForest}. Only a subset of the entire training data is used to train individual trees. The subset of the training data and input features are selected randomly. Random forests can be used for both classification and regression. Random forests can solve the decision trees' problem of overfitting as they use an ensemble of decision trees.

Random forests are known for their higher accuracy, however they are not very interpretable \cite{caruana2006empirical}. More information on random forests can be found in \cite {james2013introduction,breiman2001random}.

\subsubsection{Neural Networks}

Neural networks try to mimic the behavior of human brains. Similar to neurons in brain, neural networks have nodes that perform basic computations using activation functions. Mostly used activation functions are binary and sigmoid ones. Nodes of the neural network connect to each other using weighted edges. Neural networks are composed of three different layers: input layer (neurons in this layer get input from outer world), output layer (neurons in this layer generate predicted output response) and hidden layer(this layer performs useful transformations on input data and feed it to the output layer). There can be any number of hidden layers in a neural network. Figure \ref{fig:NNet} shows an example of a basic neural network, with only one hidden layer. This type of neural networks in which connections between layers move in forward direction only are known as feed-forward neural networks. Neural networks can have variable number of nodes and hidden layers.

Different architectures of neural networks like perceptron, multilayer perceptron, recurrent neural network and boltzman machine network have been proposed. Neural networks learn by configuring the weights of edges and thresholds of activation functions iteratively to achieve the required results. There are two main learning algorithms to train neural networks:
\textit{Gradient Descent:} In this learning algorithm the weights of the NN are modified to minimize the difference between the actual outputs and the predicted outputs. 
\textit{Back propagation:} In this learning algorithm the dissimilarity of the predicted and actual outputs is computed at the output layer and transferred to the input layer using hidden layer.

Neural networks find their use in both regression and classification problems. Neural networks are known for their good accuracy. More information on neural networks can be found in \cite{friedman2001elements,lippmann1987introduction}

\subsection{Un-Supervised Learning}
The process of training machine learning models given only sets of input vectors and no output/target vectors is called Un-supervised learning \cite{christopher2016pattern}. In other words, the training data is not labelled in case of Unsupervised learning.  This type of learning is used to attain a better understanding of the given data by identifying existing patterns in the data.

\subsubsection{Prinicpal Component Analysis}

Principal component analysis (PCA) is largely used to reduce the number of dimensions (dimensionality reduction). It allows to understand exisintg patterns in given data by using small number of "representative" variables (called principal components) from a larger number of given "correlated" variables. Basically, PCA transforms the given data to another space such that there is maximum variance among variables in new space. The components which have least variance are discarded as they do not contain much information. More information on PCA can be found in \cite{jolliffe2002principal}.

\subsubsection{Clustering}

Another popular un-supervised learning technique is clustering i.e. finding groups of data in given unlabelled data-set. Following are two main types of clustering algorithms:


\textit{K-Means Clustering:}
This algorithm starts with specification of required clusters K. Random data points are chosen as centoids of K clusters. The algorithm then identifies the points nearest to the centroid by using some distance measure, calculates mean of all points and assign a new centre to the cluster. The algorithm keeps on identifying closest points and calculate new centres untill a convergence condition is met. More information on K-means clustering can be found in \cite{james2013introduction,friedman2001elements}.


\textit{Heirarchical clustering:}
Heirarchical clustering does not require the specification of total number of clusters in advance. The algorithm builds a data binary tree that repeatedly combines similar data points. The mostly used form of heirachical clustering known as agglomerative clustering is performed as follows: Each single data point forms its own group. Two closest groups are combined iteratively, untill the time when all data points are contained in a single group/cluster. 

More information on heirarchical clustering can be found in \cite{james2013introduction,friedman2001elements}.


\subsubsection{Anomaly Detection}
Anamoly detection algorithms is a class of algorithms that identify data points with abnormal behavior in given data set. Many of the anamoly detection algorithms are unsupervised but they can be supervised and semisupervised as well \cite{chandola2009anomaly}.

\subsection{Other Types of ML Techniques}
\subsubsection{Reinforcement Learning}
Reinforcement learning is based on an "agent" attached/associated to an "environement" \cite{kaelbling1996reinforcement}. Agent decides to take certain actions depending on the state of the environment. The changes in the state because of agent's actions are fed back to the agent using reinforcement signal. Depending on the consequences of earlier decision agent receives a \textit{reward} or a \textit{penalty}. Agent should always take actions that try to increase the overall reward \cite{kaelbling1996reinforcement}, which it tries to learn by trial and error using different algorithms.

Q-learning algorithm is one of the largely used Reinforcement Learning (RL) algorithms. Interested readers can read more about reinforcement learning in \cite{sutton1998reinforcement}.

\subsubsection{Heuristic Algorithms}
These type of ML algorithms use rules or heuristics while making decisions. They work well if the solution to a problem is expensive \cite{klaine2017survey}. One of their famous types is Genetic Alorithms. These algorithms take their inspiration from nature. They use a process similar to evolution to find the best working solution to a problem. More information on genetic algorithms can be found in \cite{srinivas1994genetic}.


\subsection{Techniques Related to Model Assessment}
There are various techniques used to assess a machine learning model. Following is a description of few of them:

\subsubsection{Validation Set Approach}
This method involves partitioning the given data into two halves: training and validation/test sets. The training data set us used to train the Machine learning model. The trained model is then used for prediction of outputs using the available test set data. The error in test data is usually estimated through MSE (mean squared error). A potential problem with this approach is that the test error can have high variations \cite{james2013introduction}.

\subsubsection{Leave-One-Out Cross-Validation}

Consider that there are n total observations, then in LOOCV one observation is excluded and the model is trained on the n-1 observations. The excluded observation is used to caluclate test error. All the observations are included in the test set one by one and as a result n test errors are calculated by using training on other observations. LOOCV has less bias and randomness compared to validation set approach \cite{james2013introduction}. This method can take long time to process if the number of observations is large. 

\subsubsection{k-fold Cross-Validation}
In this apprach the available data is randomly divided into k partitions/groups of equal size. The machine learning model under consideration is trained using k-1 groups of data and then 
tested on kth group of data by calculating MSE. The aforementioned process is then replicated k times using new/different validation data sets. The overall error is estimated by calculating average of k MSE values. This method is less expensive comapred to LOOCV \cite{james2013introduction}.

\subsubsection{Bootstrapping (Bootstrap aggregating)}
\label{sec:bst}
Assuming that there is an n-sized training set called D, bagging is a process of creating m training sets $D_i$ each of size n'. These samples (also called bootstrap samples) are taken from D "uniformly with replacement". This means some data points can be repeated in $D_i$. If n'=n and n is large, $D_i$ will have nearly 63\% of unique samples of D and the others will be duplicates \cite{strickland2015data}. These m samples are used to train m ML models whose outputs are averaged to combine all of them. Bootstrapping can reduce the variance in results, but results can be more biased \cite{james2013introduction}.

\subsection{Techniques Related to Fitting of ML Models}
There are various techniques used to train a ML model. Some of them are discussed below: 

\subsubsection{Subset Selection}
This involves using only a subset of the available predictors/inputs/features to fit the ML model \cite{james2013introduction}.

\subsubsection{Best subset selection}
In this method, every combination of input features is used to train the ML model and the best combination is identified. This method is resource intensive \cite{james2013introduction}. 

\subsubsection{Step-wise selection}
Step-wise selection is more efficient way for subset selection compared to best subset selection. Forward step-wise selection starts with no predictors and keeps on adding predictors one by one untill all predictors are included in the model. At every step, the predictor that leads to greatest improvement in the model is selected. Backward step-wise selection starts with a ML model with all predictors/input varibales and removes the predictors with least benefit one by one \cite{james2013introduction}. Often hybrid versions of these two models are used. 

\subsubsection{Shrinkage methods}
To reduce variance in a regression model and avoid overfitting, often the model is trained with constraining or regularization of regression cofficient estimates. These constraints move the estimates close to zero. Such methods are known as shrinkage methods. 

\subsubsection{Ridge Regression}
In ridge regression the "coefficient estimates" are the values that minimize the expression in (9) \cite{james2013introduction}.


\begin{equation}
\sum_{i=1}^{N}( y_i-\beta_0-\sum_{j=1}^{p}\beta_jx_{ij})^2 + \lambda\sum_{j=1}^{p}{\beta_j}^2
\end{equation}

Here $\lambda$ is a "tuning parameter", p is the number of features and N is the number of data points. Ridge regression (like least sqaures) tries to find coefficients that make "RSS small" which is accounted by first term in (9). The second term in (9) is known as shrinkage penalty which controls the effect of two terms of equation "on the regression cofficient estimates". This is small when $\beta$ cofficients are close to zero \cite{james2013introduction}.  

\subsubsection{Lasso Regression}


Lasso regression is an alternative to ridge regression. It tries to find coefficients that minimize (10) \cite{james2013introduction}.

\begin{equation}
\sum_{i=1}^{N}( y_i-\beta_0-\sum_{j=1}^{p}\beta_jx_{ij})^2 + \lambda\sum_{j=1}^{p}{|\beta_j|}
\end{equation}

The difference in ridge and lasso regression is in the second term, where lasso regression has $\beta$ coefficients of ridge regression with absolute signs. Lasso regression not only punishes high values of $\beta$ like ridge regression, but it makes them zero if they seem irrelevant. This gives lasso regression its variable selection property and makes it more interpretable \cite{james2013introduction}.

\end{document}